\documentclass{INTERSPEECH2023}

\usepackage{booktabs}

\interspeechcameraready

\title{Can ChatGPT Detect Intent? Evaluating Large Language Models \\ for Spoken Language Understanding}
\name{Mutian He$^{1,2}$, Philip N. Garner$^1$}
\address{
  $^1$ Idiap Research Institute, Martigny, Switzerland \\
  $^2$ Ecole Polytechnique Fédérale de Lausanne, Switzerland
\email{\{mutian.he,phil.garner\}@idiap.ch}}

\begin{document}

\maketitle
 
\begin{abstract}
Recently, large pretrained language models have demonstrated strong language understanding capabilities. This is particularly reflected in their zero-shot and in-context learning abilities on downstream tasks through prompting. To assess their impact on spoken language understanding (SLU), we evaluate several such models like ChatGPT and OPT of different sizes on multiple benchmarks.
We verify the emergent ability unique to the largest models as they can reach intent classification accuracy close to that of supervised models with zero or few shots on various languages given oracle transcripts. By contrast, the results for smaller models fitting a single GPU fall far behind. We note that the error cases often arise from the annotation scheme of the dataset; responses from ChatGPT are still reasonable. We show, however, that the model is worse at slot filling, and its performance is sensitive to ASR errors, suggesting serious challenges for the application of those textual models on SLU.
\end{abstract}

\section{Introduction}

Gigantic pretrained language models like GPT3 with 175B parameters trained on 45TB texts have demonstrated surprisingly strong performance on various downstream language tasks with little or no data 
\cite{DBLP:conf/nips/BrownMRSKDNSSAA20}. Since then, GPT3 has evolved into GPT3.5 through pretraining on code as in Codex \cite{DBLP:journals/corr/abs-2107-03374} which powers GitHub Copilot, as well as through instruction fine-tuning that aligns the model's responses given instructions with human expectations using reinforcement learning, known as InstructGPT \cite{DBLP:journals/corr/abs-2203-02155}. When further combined with fine-tuning on dialogues in a similar way, the resulting model, ChatGPT, has gained great popularity since its release in late 2022, displaying highly human-like language understanding and generation capabilities \cite{kung2023performance,DBLP:journals/corr/abs-2302-04023,DBLP:journals/corr/abs-2302-06476}, and has become the core of a number of AI-powered applications. Combined with the ability to utilize tools with external APIs as in Toolformer \cite{DBLP:journals/corr/abs-2302-04761} as well as conducting web search as in WebGPT \cite{DBLP:journals/corr/abs-2112-09332} and New Bing, a competent and versatile AI assistant has taken shape. Therefore, a question arises: is the model capable of conducting spoken language understanding (SLU) tasks like current voice assistants?

Current SLU approaches are substantially different from how we use those GPT3-based models. Traditionally, SLU is carried out using a cascaded pipeline, which includes an automatic speech recognition (ASR) module taking audio as inputs, and a natural language understanding (NLU) module working on ASR transcripts, hypotheses, or lattice to predict labels for tasks like intent classification (IC) and slot filling (SF) \cite{DBLP:conf/asru/Mori07,DBLP:conf/ijcai/QinXC021}. Recently, end-to-end approaches that directly predict labels from speech \cite{DBLP:conf/icassp/SerdyukWFKLB18,DBLP:conf/slt/HaghaniNBCGMPQW18,DBLP:conf/interspeech/SaxonCMM21} become more popular, and pretrained language and speech models are also introduced into SLU \cite{DBLP:journals/corr/abs-2111-02735,DBLP:conf/icassp/AroraDDCUPZKGYV22,DBLP:conf/icassp/SeoKL22}. Additionally, there are works focused on low-resource or few-shot textual IC/SF \cite{DBLP:conf/emnlp/YazdaniH15,DBLP:conf/interspeech/FerreiraJL15,DBLP:conf/acl/HouCLZLLL20,DBLP:conf/emnlp/WuSJ21,DBLP:conf/interspeech/PengZZG21}.

However, those methods are based on the paradigm of supervised training or fine-tuning with a set of possibly large-scale training data. In contrast, considering the difficulty of fine-tuning the whole GPT3 model, recent NLP research highlights a different scheme, namely \textbf{prompting} \cite{DBLP:journals/csur/LiuYFJHN23}: given a fixed textual description of the task known as a \textit{prompt} without any training data, the language model may correctly carry out the task. Furthermore, the \textbf{in-context learning} approach adds a few paired examples in the textual prompt to further direct the model towards the desired outputs. Such methods are different from traditional zero or few-shot learning in which the model parameters are fixed. It appears to be clumsy and may perform worse than fine-tuned smaller models like T5-11B at the beginning \cite{DBLP:conf/emnlp/LesterAC21}. Additionally, larger models were believed to be unscalable to reach the desired performance given the costs \cite{DBLP:conf/nips/BrownMRSKDNSSAA20,DBLP:journals/corr/abs-2001-08361,DBLP:journals/corr/abs-2110-14168}, However, recent explorations reveal the \textbf{emergent abilities} of larger models like GPT3-175B and PaLM 540B \cite{DBLP:journals/corr/abs-2204-02311}: prompting shows low or even close-to-random performance on multiple tasks until a certain scale of the model where a breakthrough emerges \cite{DBLP:journals/corr/abs-2206-04615,DBLP:journals/corr/abs-2206-07682}. This breakthrough enables chain-of-thought prompting to surpass the smaller models fine-tuned on rich data \cite{weichain,DBLP:journals/corr/abs-2210-00720}, allows reasoning using internal knowledge with results comparable to external knowledge retrievers \cite{DBLP:journals/corr/abs-2209-10063}, and leads to better robustness and generalization \cite{DBLP:journals/corr/abs-2210-00720,DBLP:journals/corr/abs-2210-09150}.

There have been several works on SLU employing prompts, such as fine-tuning pretrained models like T5 aided by prompts \cite{DBLP:conf/coling/WuWZCZ22,DBLP:journals/corr/abs-2210-03337}, fine-tuning embeddings prepended to the inputs known as continuous prompts \cite{DBLP:conf/interspeech/ChangT0L22,DBLP:journals/corr/abs-2303-00733}, and end-to-end SLU by in-context learning on GPT2 with a fine-tuned audio encoder \cite{DBLP:conf/interspeech/GaoNQZCH22}. They are nevertheless distinct from the current prompting and in-context learning scheme, and have not approached the regime of emergent abilities. Hence the potential and limitations of this new type of method on SLU remain unexplored. Therefore we endeavor to undertake it by designing prompts and evaluating these models on multiple SLU benchmarks, including SLURP \cite{DBLP:conf/emnlp/BastianelliVSR20} and the multilingual MINDS-14 \cite{DBLP:conf/emnlp/GerzSKMLSMWV21}. Since these models take textual inputs, beside oracle transcripts, we also use ASR transcripts from Whisper \cite{DBLP:journals/corr/abs-2212-04356}, which embodies a pipeline upon completely off-the-shelf pretrained models. Furthermore, we compare smaller models that can easily run on a common GPU, namely GPT2 \cite{radford2019language} and several OPT models \cite{DBLP:journals/corr/abs-2205-01068}.

As a result, we discover that the largest GPT3.5 and ChatGPT models achieve high performance on intent classification under zero-shot or few-shot in-context scenarios that are close or even better than models fine-tuned on the whole dataset, when given the oracle transcripts. This is unique to those large models as the smaller GPT2, OPT and GPT3.5 Curie models have much lower performance and are entirely ineffective under zero-shot cases. Even in the cases where the predictions differ from the labels, the predictions are mostly reasonable, often due to the ambiguity of the sentence. This raises the question that the tasks to predict intermediate IC/SF labels might not sufficiently reveal the potential of the model. However, for the slot filling task with a more complicated task definition, the performance is much worse. Additionally, the accuracy drops significantly when using ASR transcripts. We show that the models have limited awareness of word pronunciations and possible ASR errors, which poses challenges for directly deploying those models for real-world SLU. To facilitate reproduction, relevant resources and prompts are available at \url{https://mutiann.github.io/papers/ChatGPT_SLU/}.

\section{Methods}

\begin{figure}[t]
  \centering
  \includegraphics[width=\linewidth]{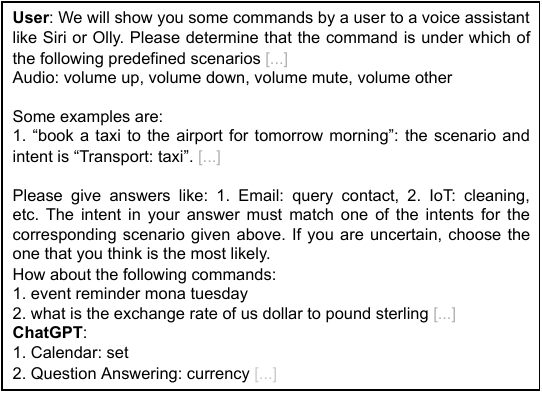}
  \caption{An example of ChatGPT doing SLURP intent classification in a conversation. The list of options, questions, and answers are partially omitted for brevity.}
  \label{fig:prompt}
\vspace{-20pt}
\end{figure}

We design prompts to be given to ChatGPT for the intent classification and slot filling tasks as in Figure~\ref{fig:prompt}. We begin by explaining the background of the task, and then provide options, like possible intents. We rewrite the names of the scenarios and actions in natural language, for example, with the underlines in the name removed. For in-context learning cases, several examples are further appended. Following by instructions about the answer format are the questions. Several questions are asked in a batch for better efficiency. For instance, in the case of ChatGPT on SLURP, we ask 45 questions in a conversation, with the first 5 questions appended to the instructions, and the remaining 40 questions split into 2 rounds to be asked. We then collect results from each line of the answer that corresponds with each question by text matching. If the answer could not be identified, we retry up to 3 times. As for ChatGPT, we use the \textit{legacy} (a.k.a. \textit{default}, codenamed \textit{text-davinci-002-render-paid}) model available on the webpage, which gives slightly different results compared to the faster \textit{default} (a.k.a. \textit{turbo}, codenamed \textit{text-davinci-002-render-sha}) model. For slot filling, we further explain the meaning of each entity type. While as for GPT3.5, we adopt the largest \textit{text-davinci-003} model as well as the smaller \textit{text-curie-001} model\footnote{The sizes of Davinci and Curie are estimated to be comparable to GPT3 175B and 6.7B respectively, according to Eleuther AI as in \url{https://blog.eleuther.ai/gpt3-model-sizes/}}, using a slightly different prompt to fit the task of text completion. Since GPT3.5 is called by the OpenAI API, more customization is allowed, including logit bias to ensure that only tokens that may appear in the answer could be generated. While as for smaller models, we use GPT2 large (774M) \cite{radford2019language}, the widely-used predecessor, as well as OPT, an open-sourced reproduction of GPT3 with various sizes available \cite{DBLP:journals/corr/abs-2205-01068}. We picked the 1.3B, 2.7B, and 6.7B versions that could be easily run on a single GPU, though we have to use half-precision inference on the 6.7B one. We performed generation with an 8-beam search, using a prompt similar to GPT3.5 but perform text completion with one question at a time. Whisper-large is chosen for performing ASR \cite{DBLP:journals/corr/abs-2212-04356}. 

The models are evaluated on the test split of two different datasets: SLURP is a large-scale IC and SF dataset of commands to voice assistants with over 141k samples annotated with 60 different intents formulated as scenario-action pairs, as well as 56 types of entities or slots. There have been limited evaluations on a task with so many different types, as current works are mostly focused on tasks with a small label space (e.g. positive/neutral/negative in sentiment analysis), though lower performance has been found on named entity recognition (NER) \cite{DBLP:journals/corr/abs-2302-06476} and dialogue state tracking \cite{DBLP:journals/corr/abs-2302-04023} with a smaller label space but share some similarity with this task.
MINDS-14 is a smaller banking scenario IC dataset of 14 intents in 14 different languages, for which we use the XTREME-S split \cite{DBLP:conf/interspeech/ConneauBZMPLCJR22}. Evaluations are performed on the test split, while in-context training examples are picked from the training or validation split.

\section{Experiments}
\subsection{Oracle texts}
\begin{table}[t]
  \caption{Accuracy for various GPT3-based models on SLURP intent classification task with zero or few examples.}
  \label{tab:gpt_slurp}
  \centering
  \begin{tabular}{lcccc}
  \toprule
\#Examples                             & 0       & 10      & 20      & 30      \\ \midrule
GPT3.5                       & 72.86\% & 74.55\% & 77.27\% & 77.44\% \\
~~~~w/ bias       & 75.86\% & 75.59\% & 78.31\% & 77.87\% \\
~~~~Curie w/ bias & 5.01\%  & 4.91\%  & 3.80\%  & 3.77\%  \\
ChatGPT                    & 79.25\% & 80.33\% & \textbf{83.93\%} & 80.16\% \\
~~~~Turbo ver.                     & 78.98\% & 80.03\% & 81.78\% & 79.62\%\\ \bottomrule
  \end{tabular}
  \vspace{-10pt}
\end{table}

We first evaluate the models from OpenAI using oracle transcripts in SLURP for intent classification. As shown in Table~\ref{tab:gpt_slurp}, ChatGPT achieves an accuracy of 79.25\% with zero shot, and 83.93\% with only 20 examples. Despite there being 60 different types of intents in SLURP, ChatGPT achieves the result even without examples in many of the categories. The result is on par with, or not far from, many supervised NLU models trained on the full oracle transcripts dataset, like SF-ID with 82.25\%, HerMiT with 84.84\% \cite{DBLP:conf/emnlp/BastianelliVSR20}, and conformer deliberation with 89.0\% \cite{DBLP:conf/interspeech/AroraDCYB022}, or end-to-end methods like fine-tuning HuBERT with 89.38\% \cite{DBLP:journals/corr/abs-2111-02735}. While more training examples seem to have limited or no improvements, as shown by the results on 30 examples. This could be possibly due to the bloated length of the prompt. When we use a more verbose version of the prompt with 20 examples but with 1260 tokens, similar to the 30-example one, the accuracy drops to 81.41\%. 
ChatGPT gives slightly better performance compared to GPT3.5 even with token bias, possibly thanks to its additional dialogue fine-tuning, while the faster Turbo version shows slightly inferior results. 

\begin{table}[t]
  \caption{Accuracy for smaller language models on SLURP intent classification task with zero or few examples. A 30-example prompt is too long for GPT2 with a 1024 token window.}
  \label{tab:lm_slurp}
  \centering
  \begin{tabular}{lcccc}
  \toprule
\#Examples                             & 0       & 10      & 20      & 30      \\ \midrule
GPT2 (774M) & 6.66\% & 8.88\%  & 8.31\%  & -       \\
OPT-1.3B          & 5.58\% & 10.69\% & 17.85\% & 17.01\% \\
OPT-2.7B          & 7.06\% & 28.65\% & 26.66\% & 36.97\% \\
OPT-6.7B  & 4.37\% & 28.18\% & 35.14\% & \textbf{42.40\%} \\
\bottomrule
  \end{tabular}
  
\end{table}

The results obtained with smaller models are presented in Table~\ref{tab:lm_slurp} in addition to the Curie results above. It is evident that the performance falls far behind the presumed 175B GPT3.5 and ChatGPT models. The model size is apparently the most critical factor affecting the performance, while the number of examples also makes a difference. Relatively larger (e.g. 6.7B) models can better leverage the increased examples, while none of the zero-shot experiments work. This is in stark contrast to the largest models where the number of examples has limited impact and the zero-shot cases also show good results. This suggests that the largest models possess sufficient capabilities through pretraining to comprehend the task explanation and the input question without additional training examples. The role of the training examples is more to guide the model to align with the requirements of the task that are not elaborated in the task description. In contrast, in-context examples are more important for smaller models that lack sufficient internal knowledge. 

We then evaluate the models on four different languages from MINDS-14 selected based on their proportion in the GPT3 training data: Among 118 languages in the data, 92.6\% are in English by word count, while the 2nd-ranked French accounts for 1.8\%. Other languages are exceptionally sparse: Polish (ranked 10th) for 0.16\%, and Korean (28th) for 0.017\%. Nevertheless, ChatGPT generalizes to all of these languages with similar or better performance compared to the supervised LaBSE \cite{DBLP:conf/acl/FengYCA022} reported in \cite{DBLP:conf/emnlp/GerzSKMLSMWV21}, though using a different split. The model can almost perfectly solve the task on English and the rather low-resource French with zero shot, as well as Polish using only 14 examples or 1 shot per category to align with the task. Even on Korean with extremely sparse training data, the zero-shot and one-shot results are still satisfactory. This demonstrates that the model enjoys inherent multilinguality to generalize its strong language understanding capability to various languages.

\begin{table}[t]
  \caption{Accuracy for intent classification on different languages from MINDS-14 with ChatGPT, compared to the supervised textual model using full data.}
  \label{tab:chatgpt_minds14}
  \centering
  \begin{tabular}{lcccc}
  \toprule
 & en-US   & fr-FR   & pl-PL   & ko-KR   \\ \midrule
0-shot          & 95.4\% & 97.4\% &  90.0\%       & 89.2\% \\
1-shot         & 97.9\% & 99.3\% &   96.1\%      & 90.5\% \\ \midrule
LaBSE \cite{DBLP:conf/emnlp/GerzSKMLSMWV21} & 95.1\% & 93.1\% & 89.2\% & 91.4\% \\ \bottomrule
  \end{tabular}
  \vspace{-15pt}
\end{table}

\begin{table}[t]
  \caption{F1 score for slot filling with ChatGPT with different number of examples.}
  \label{tab:chatgpt_slot}
  \centering
  \begin{tabular}{lccc}
  \toprule
\#Examples &  10      & 20      & 30    \\  \midrule 
ChatGPT & 12.03\% & 13.00\% & 13.35\% \\ \bottomrule
  \end{tabular}
  \vspace{0pt}
  
\end{table}

However, when it comes to the slot filling task with a more complicated task definition, the situation is different. As in Table~\ref{tab:chatgpt_slot}, the F1 score is poor and much lower than the models like HerMiT with 78.19\% F1 \cite{DBLP:conf/emnlp/BastianelliVSR20}. Therefore, the approach to better leveraging ChatGPT on this task remains to be explored.

\subsection{ASR transcription}
\begin{table}[t]
  \caption{Accuracy for intent classification on SLURP with different number of examples using ASR transcripts.}
  \label{tab:chatgpt_asr}
  \centering
\begin{tabular}{lccc}
  \toprule
\#Examples & 10      & 20      & 30      \\ \midrule
GPT3.5           & 64.78\% & 68.95\% & 68.64\% \\
ChatGPT        & 72.89\% & 73.96\% & 72.85\%        \\ \bottomrule
  \end{tabular}
  \vspace{-10pt}
\end{table}

We then evaluate the model on SLURP intent classification using ASR transcripts, which reflects the real SLU scenario. The SLURP recordings are transcribed with 16.7\% test WER with the off-the-shelf Whisper, which is acceptable, considering that the audio is often noisy and an XLSR-based ASR system adapted to the SLURP data reports 15.5\% test WER \cite{DBLP:journals/corr/abs-2212-08489}. To our disappointment, the models suffer from significant performance drop processing ASR transcripts as in Table~\ref{tab:chatgpt_asr}, even though the model is prompted to account for ASR errors.

Considering that GPT3 is capable of correcting language errors, we are inspired by the chain-of-thought prompts \cite{weichain} to first instruct the model to try to correct the ASR errors, and then determine the scenario and action. Such a prompt with 5 examples brings 67.15\% accuracy on GPT3.5, which alleviates the issue with a 2.4\% improvement compared to the 10-example prompt, although it is still far from the oracle transcript case and leads to bloated prompt. We also attempt to directly instruct GPT3.5 to correct ASR errors, characterized by the replacement of words with similar pronunciation. Given 1000 error cases, WER increases from 25.6\% to 32.4\% after the correction. In most cases the sentences are more fluent, but ASR errors remain. Hence we hypothesize that due to the nature of training on textual inputs with wordpiece tokenization, the model has limited awareness of phonetics, as implied by their high error rates on the IPA  task \cite{DBLP:journals/corr/abs-2206-04615}. We further verify that by asking ChatGPT the names of 50 well-known places with their Romanized transliteration in Chinese and Japanese, e.g., ``A city is called Nyūyōku in Japanese. What's its name in English?'', which can be easily guessed by a person knowing the common pronunciation of Latin letters. However, ChatGPT can correctly answer only 46\% from Japanese and 10\% from Chinese \footnote{Later we experimented with the more recent GPT4 model, and it achieves 100\% accuracy on this Chinese test. It also appears to have stronger capability for correcting ASR mistakes, but errors are still often. Therefore a stronger textual model may partially alleviate this limitation, though the extent remains unclear and is left for future investigation.}, which further confirms its deficiency in phonetic knowledge.

\section{Discussions}
\label{sec:error}

\begin{figure}[t]
  \centering
  \includegraphics[width=\linewidth]{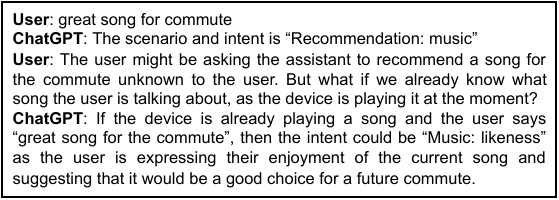}
  \vspace{-15pt}
  \caption{An example for ChatGPT giving an answer different from the label (“Music: likeness”) on an ambiguous command. While ChatGPT is able to determine the answer that matches the label under an alternative situation.}
  \label{fig:conversation}
  \vspace{-20pt}
\end{figure}

\begin{table*}[t]
  \caption{Examples of “errors” made by ChatGPT on SLURP intent classification and slot filling. Many of the predictions are also reasonable, but different from the labels in the dataset.}
  \label{tab:chatgpt_errors}
  \centering
\begin{tabular}{p{6.5cm}p{4.3cm}p{4.8cm}}
  \toprule
Command                                                                         & Label                          & ChatGPT Prediction                     \\ \midrule
Is there a groomer in town for cats only?                                        & General: quirky                & Recommendation: locations \vspace{2px}\\ 
In how many hours will it be midnight in London England.                       & Date/time: query             & Date/time: convert          \vspace{1px}\\
Please tell me news related to the stock market.                               & Question answering: stock    & News: query               \vspace{2px}\\
Olly I need a drink.                                                           & IoT: coffee                    & General: quirky             \vspace{2px}\\
Can you give me the details of upcoming annual function on twenty sixth march? & General: quirky                & Recommendation: events     \vspace{1px}\\
How long does it take to make vegetable lasagna?                               & Cooking: recipe                & Cooking: query              \vspace{2px}\\
Give details of rock sand.                                                     & Question answering: definition & Question answering: factoid
\\ \midrule
Event reminder Mona Tuesday.                                & Event name: Mona; \newline Date: Tuesday                            & Alarm type: reminder; Person: Mona; Date: Tuesday                                  \vspace{1px}\\
Exchange rate of US dollar to pound sterling.             & News topic: exchange rate of US dollar to pound sterling & Definition word: exchange rate; \newline Currency name: US dollar; Currency name: pound sterling \vspace{2px} \\ 
Take out the milk from the shopping list.                   & List name: shopping                                        & List name: shopping; Ingredient:   milk                                               \vspace{1px} \\
Increase the brightness of the lights.                      & -                                                          & Change amount: brightness; Device type: lights                                    \vspace{2px} \\
Please send a mail to my friend Divya how are you.        & Relation: friend; Person: Divya                            & Email address: Divya; Person: friend                      \vspace{2px} \\
Please scan my social media and tell me what's happening. & -                                                          & Media type: social media; \newline Definition word: what's happening                               \\  \bottomrule
  \end{tabular}
  \vspace{-14pt}
\end{table*}

Experiments have demonstrated the capability of the model, though many errors are found. However, it is worthwhile to look into the error cases to determine if ChatGPT really failed to understand those text, especially considering that SLURP labels are relatively noisy \cite{DBLP:journals/corr/abs-2212-08489}. Therefore, we check 100 error cases in the 20-shot ChatGPT experiments, and find that 79\% of them are not errors in the strict sense, as exemplified in Table~\ref{tab:chatgpt_errors}. Some of the “errors” occur when the input sentence is ambiguous and open to multiple different interpretations. An example is shown in Figure~\ref{fig:conversation}, in which case we find that ChatGPT could actually provide the correct answer and explain the answer if we provide clarification that is not given in the command. There are also some examples for which ChatGPT gives a more accurate answer than the label. While in many other cases the issues arise due to the ambiguity of the categories, which are often specified by the annotation scheme of the SLURP dataset. For example, asking for news about stock is labeled as “Question answering: stock,” not “News: query.” This is a fundamental weakness of prompting: the instructions, examples, and label names in the prompt must fully reveal the goal of the task if it is not typically observed in the textual corpus, which is difficult when the task or labeling specification is complicated and subtle, even if the task itself is straightforward.
Such annotation specifications could be captured by a supervised model given the whole training set, but impossible for a zero-shot or in-context model.

The phenomenon is more pronounced in the slot filling case where multiple entities with 56 different types could be extracted from a single command. The meanings of the labels often overlap with each other, and the annotations tailored to the specific scenario of the dataset. For example, normal apps are extracted as “app name,” and companies as “business name”, but Uber as “transport agency” and Northern Rail as “transport name,” which would baffle a zero/few-shot model as in Table~\ref{tab:chatgpt_errors}. The harm of such counter-intuitive labelling scheme also matches the observations in \cite{mckenzie2022inverse}. This kind of misalignment has more impact in such a tagging task as tagging decisions need to be made on every individual word, which may explain the observed worse performance on similar tasks like NER and dialogue state tracking \cite{DBLP:journals/corr/abs-2302-04023,DBLP:journals/corr/abs-2302-06476}. We also find that ChatGPT may be biased towards patterns in the limited examples. For example, with “tell me about morel mushrooms” labeled as “Definition word: morel mushroom” in a training example, ChatGPT tends to label “Definition word” on other sentences like “tell me what's happening.”  All these examples illustrate the difficulties of in-context learning on a task with a complicated task goal.

In the evaluations conducted above, we direct the model to fulfill requirements of the dataset, which are intermediate tasks in a pipeline. The predictions are then utilized to run external functions (e.g., sending an email, answering a question, telling a joke) available in the specific annotated scenario using certain rules. However, ChatGPT has demonstrated its versatility as a chatbot that can communicate with users without being bound by these objectives. For instance, the model often appears to “understand” the commands in the “error” cases above, and may directly request for clarification when the input can not be understood, such as when an ASR error happens. When equipped with the ability to call external APIs \cite{DBLP:journals/corr/abs-2302-04761, DBLP:journals/corr/abs-2112-09332}, New Bing can read returned results and reason upon them to determine the final response to users' requests. Additional instructions regarding the functions can be also fed to the model via some system prompt in multiple rounds of conversation along with user inputs. In this way, the model's ability on such intermediate tasks may underestimate its actual capability as an end-to-end assistant.

\section{Conclusions}

In this paper we evaluate large language models on SLU tasks with prompting. We find that they reach results comparable to fine-tuning on intent classification of multiple languages, but not on slot filling, and being sensitive to ASR errors. Future work will be focused on better prompting and incorporating phonetic knowledge for ASR-robust or end-to-end systems.

\section{Acknowledgement}
This work received funding under project SteADI, Swiss National Science Foundation grant 197479.

\bibliographystyle{IEEEtran}
\bibliography{mybib}

\end{document}